\documentclass[conference]{IEEEtran}
\IEEEoverridecommandlockouts
\usepackage{cite}
\usepackage[T1]{fontenc}
\usepackage{color}
\usepackage{hyperref}
\hypersetup{
    colorlinks=true, 
    citecolor=green,  
    linkcolor=red,  
    urlcolor=blue    
}
\usepackage{amsmath,amssymb,amsfonts}
\usepackage{algorithm}
\usepackage{algpseudocode}
\usepackage{amsmath}
\usepackage{amssymb}
\usepackage{subcaption}
\usepackage{graphicx}
\usepackage{lipsum}
\usepackage{amsmath,amssymb,amsfonts}
\usepackage{graphicx}
\usepackage{xcolor}
\usepackage{multirow}
\usepackage{multicol}
\usepackage{booktabs,caption}
\usepackage{color}

\usepackage{comicneue}
\begin{document}

\title{{\comicneue\textbf{MGFace}}: Mask-Gated Face Matching via Conditional Similarity Routing}

\author{\IEEEauthorblockN{Huy Che\textsuperscript{1, 2}, Hoang-Minh Trinh\textsuperscript{1, 2}, Dinh-Duy Phan\textsuperscript{1, 2} and Duc-Lung Vu\textsuperscript{1, 2,*}}
\IEEEauthorblockA{\textsuperscript{1}University of Information Technology, Ho Chi Minh City, Vietnam}
\IEEEauthorblockA{\textsuperscript{2}Vietnam National University, Ho Chi Minh City, Vietnam}
Email: huycq@uit.edu.vn, 21522347@gm.uit.edu.vn, duypd@uit.edu.vn, lungvd@uit.edu.vn
\IEEEauthorblockA{
\textsuperscript{*}Corresponding author: Duc-Lung Vu
}
}

\maketitle          

\begin{abstract}
Face identification has achieved remarkable performance under normal conditions. Yet, its accuracy often degrades significantly when query faces are partially occluded, especially by facial masks. Existing re-ranking approaches improve robustness by exploiting patch-level similarities. Still, they often rely on costly, fine-grained matching mechanisms, which limit their efficiency in large-scale retrieval scenarios. In this paper, we propose MGFace, a mask-gated face identification pipeline that predicts the mask status of a query face and conditionally routes the similarity computation accordingly. Specifically, MGFace distinguishes between masked and unmasked queries, applies global embedding matching to unmasked queries, and activates mask-aware patch-level re-ranking only for masked queries. This design focuses on reliable upper-face regions while avoiding unnecessary fine-grained computation. Experiments on the extended LFW-Mask dataset show that MGFace achieves over 80\% identification accuracy with the FaceNet backbone and over 90\% with the ArcFace backbone. Compared with a previous EMD-based re-ranking method, MGFace achieves better identification performance while reducing query time by approximately 20$\times$. These results demonstrate the effectiveness of MGFace in improving masked-face identification accuracy with low computational overhead. The source code is available at \url{https://github.com/chequanghuy/MGFace}.
\end{abstract}


\section{Introduction} \label{intro}

Face identification \cite{arcface,cosface,facenet,centerloss,transface} is a fundamental computer vision task that aims to identify a person from facial images. It has been widely deployed in real-world applications such as access control, surveillance, and intelligent attendance systems \cite{face_class_mapr}. As multimedia databases continue to grow in scale, practical face identification systems must satisfy two requirements simultaneously: they must remain accurate under challenging visual conditions and efficient enough for large-scale retrieval.

Recent deep face recognition models achieve strong performance under common variations such as pose, illumination, and age \cite{agedb,vggdataset}. However, their accuracy can degrade significantly when discriminative facial regions are occluded by masks or sunglasses \cite{emd,fastface}. masked-face identification is especially challenging because masks hide the lower face and remove important identity cues; previous work \cite{emd} reports that identification accuracy can drop from 98.41\% to 39.79\% when the query face is masked. To improve robustness, existing two-stage methods \cite{emd,fastface} first retrieve candidates using global embeddings and then refine the ranking with patch-level similarity. Although effective, these methods often incur costly local matching for all queries, even for unmasked faces where global embeddings are already reliable. Moreover, patch-level matching may include occluded lower-face regions, which can introduce unreliable similarity signals. Thus, existing re-ranking methods \cite{emd,fastface} improve masked-face retrieval but do not explicitly address when fine-grained matching should be used or where it should focus.

\begin{figure}
    \centering
    \includegraphics[width=\linewidth]{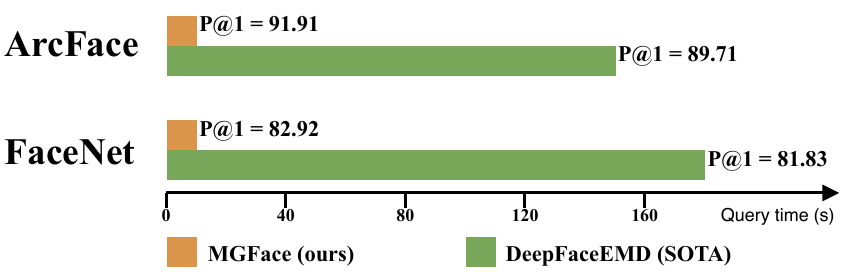}
    \caption{Comparison of query times during the evaluation phase and P@1 accuracy of MGFace and DeepFaceEMD.}
    \label{fig:highlight}
\end{figure}

In this paper, we propose MGFace, a mask-gated face identification pipeline for efficient masked-face retrieval. MGFace predicts the query mask status using a lightweight head integrated into a pretrained face recognition backbone, and then routes the query to an appropriate matching branch. Unmasked queries are matched directly with global cosine similarity, while masked queries are processed by a two-stage mask-aware branch that retrieves top-$k$ candidates and re-ranks them using patch-level similarity over reliable upper-face regions. This design applies fine-grained matching only when necessary and avoids unreliable occluded patches. 
Figure \ref{fig:highlight} illustrates the effectiveness of the proposed method, achieving higher accuracy with significantly lower query times.

Our main contributions are summarized as follows: 
\textbf{(1)} We propose MGFace, a mask-gated face identification pipeline that augments pretrained face recognition models with a lightweight mask classification head to predict whether a query face is \texttt{masked} or \texttt{unmasked}. 
\textbf{(2)} We introduce a conditional similarity routing strategy that adaptively selects the matching process according to the predicted mask status: unmasked queries are matched using global cosine similarity, while masked queries are processed by mask-aware patch-level re-ranking over reliable upper-face regions. 
\textbf{(3)} We evaluate MGFace on the extended LFW-Mask dataset with different pretrained backbones, demonstrating that it improves face identification accuracy under masked-face conditions while reducing computational cost compared with existing re-ranking-based methods.

\begin{figure*}[h]
    \centering
    \includegraphics[width=0.8\linewidth]{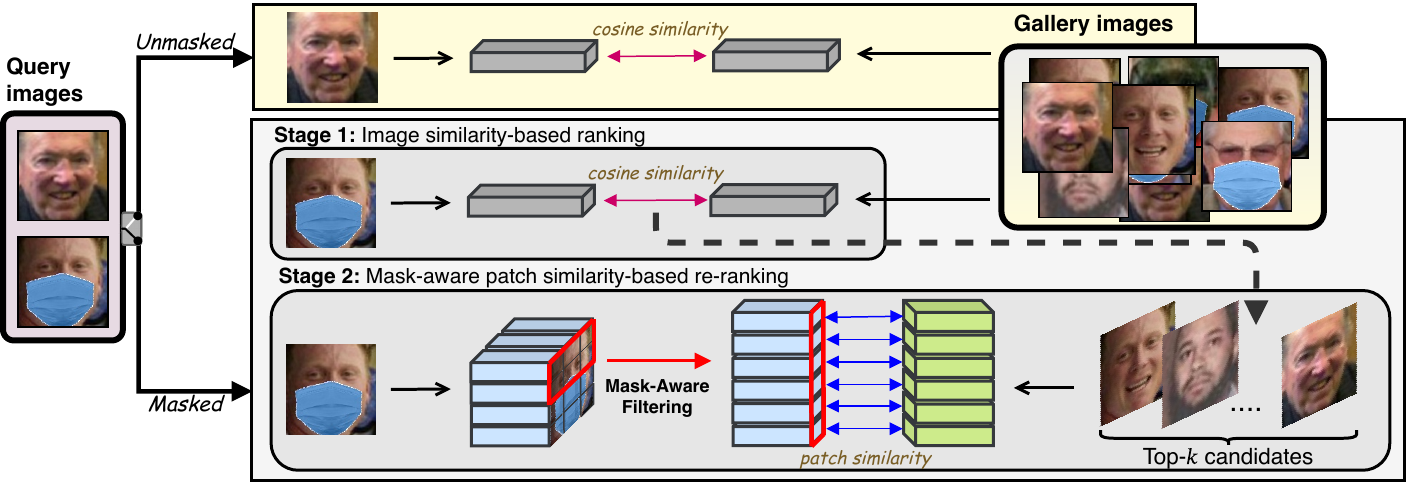}
    \caption{Overview of the proposed MGFace pipeline. Given a query face, MGFace first predicts its mask status and conditionally routes it to different matching branches. Unmasked queries are identified using \textit{global face matching}, while masked queries are processed by \textit{two-stage mask-aware matching}, which retrieves top-$k$ candidates using global similarity and re-ranks them with patch-level similarity over reliable facial regions.}
    \label{fig:pipeline}
\end{figure*}

\section{Related work}

Face identification methods commonly rely on CNN-based \cite{arcface,facenet, cosface}, Transformer-based \cite{transface}, or hybrid architectures \cite{fastface} as backbone networks to extract deep facial representations. These models are typically trained using classification-based learning \cite{arcface,cosface} or metric learning objectives \cite{facenet,centerloss}, which encourage images from the same identity to be close in the embedding space while pushing images from different identities farther apart. Although such methods achieve strong performance on in-distribution benchmarks \cite{agedb,vggdataset}, their accuracy often degrades under out-of-distribution conditions, such as masked faces and sunglasses. Facial masks are a common and challenging form of occlusion, particularly in real-world identification systems where masked queries became widespread during the COVID-19 pandemic. Consequently, face-related tasks involving masks \cite{face_detect,mask_class,counting_face}, especially masked-face identification, have attracted increasing research attention. To address masked-face identification, prior works retrain models on synthetic masked images~\cite{occlu1,occlu2}, reconstruct occluded regions~\cite{deocc1,deocc2}, or aggregate multiple features~\cite{fuse1,fuse2}. However, these methods often require extra training, may introduce unreliable information, or increase model complexity. In addition, feature aggregation methods~\cite{fuse1,fuse2} have been explored to enrich visual representations, while recent re-ranking strategies aim to improve identification under OOD scenarios~\cite{emd,fastface,kwf}. For example, DeepFace-EMD \cite{emd} exploits Earth Mover’s Distance to compare patch-level features, thereby improving robustness but substantially increasing computational cost during re-ranking. Hybrid-ViT \cite{fastface} uses Transformer features and requires additional training on masked-face data. These methods also apply fine-grained matching to all queries, causing unnecessary computation for unoccluded faces. A straightforward solution is to use an addition classifier \cite{mask_class,reidmulticlass} to select matching strategies based on query conditions. However, employing an additional standalone classifier introduces extra inference overhead. 

In contrast, our work proposes a mask-gated face identification pipeline that integrates a lightweight mask classification head into a pretrained face recognition model. This enables efficient conditional routing between global matching and mask-aware local re-ranking, improving robustness under masked-face scenarios while reducing unnecessary computation.

\section{Method}

In this section, we introduce MGFace, a mask-gated face identification pipeline that adaptively selects the matching strategy according to the mask status of the query face. Instead of applying a uniform matching procedure to all queries, MGFace first estimates whether the query face is masked and then routes it to an appropriate identification branch. For unmasked queries, the full facial region is considered reliable, and identification is performed using \textit{the global matching branch}. For masked queries, where the lower facial region may be occluded, MGFace employs \textit{the two-stage mask-aware branch}. It first retrieves the top-$k$ gallery candidates using global image-level similarity and then re-ranks these candidates by computing patch-level similarity over reliable facial regions after mask-aware patch filtering. Figure~\ref{fig:pipeline} provides an overview of the proposed mask-gated face identification pipeline.

\subsection{Extending pretrained face recognition models}

To separately process masked and unmasked face images for identity recognition, a straightforward approach is to employ an additional mask classification model alongside the face recognition model. However, such a design substantially increases inference latency, as the system must sequentially execute two separate models: one for mask-status classification and another for facial feature extraction. To address this limitation, we propose an integrated architecture built upon a pretrained face recognition model. Specifically, the pretrained face recognition backbone is frozen, and an additional classification head is attached to the backbone's convolutional layer to predict two classes: \texttt{masked} and \texttt{unmasked}. The classification branch applies global average pooling followed by fully connected layers to estimate the probability of whether a face is masked. This design enables simultaneous facial feature extraction and mask-status classification within a single forward pass, thereby reducing computational overhead compared with using two independent models. Figure \ref{fig:model} illustrates the proposed architecture. Given an input image, the proposed model produces three outputs:

\begin{figure}[h]
    \centering
    \includegraphics[width=0.8\linewidth]{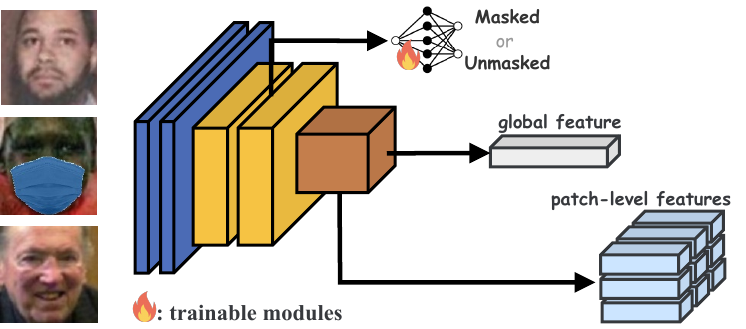}
    \caption{Proposed architecture with an classification head. The pretrained backbone extracts global and patch-level features, while a lightweight head predicts the mask status.}
    \label{fig:model}
\end{figure}

\begin{equation}
(f, P, p)=\mathcal{F}(\cdot)
\end{equation}

where $f \in \mathbb{R}^{d_f}$ is the global feature of the image, $P \in \mathbb{R}^{d_p \times N \times N}$ is the patch-level features, and $p \in \{\texttt{masked}, \texttt{unmasked}\}$ is the predicted mask status.

\subsection{Mask-Gated Face Identification}

Given a query face image $q$, MGFace represents it as $(f_q, P_q, p_q)$, where $f_q$ is the global feature, $P_q$ is the patch-level feature map, and $p_q$ is the predicted mask status. Similarly, each gallery image $g_j$ in the gallery set $\mathcal{G}=\{g_j\}_{j=1}^{n}$ is represented as $(f_j, P_j)$. Based on $p_q$, MGFace routes the query to one of two matching branches:

\begin{equation}
\mathcal{R}(q,\mathcal{G})=
\begin{cases}
\mathcal{R}_{\text{global}}(q,\mathcal{G}), & p_q=\texttt{unmasked},\\
\mathcal{R}_{\text{patch}}(q,\mathcal{G}), & p_q=\texttt{masked}.
\end{cases}
\end{equation}

Here, $\mathcal{R}_{\text{global}}(q,\mathcal{G})$ denotes the ranked list produced by global feature matching, where gallery images are sorted according to their cosine similarity with the query global feature. In contrast, $\mathcal{R}_{\text{patch}}(q,\mathcal{G})$ denotes the ranked list produced by the two-stage mask-aware matching branch. This branch first retrieves the top-$k$ candidates using global similarity and then re-ranks them using patch-level similarity computed only over reliable facial regions.

\subsubsection{Global Face Matching for Unmasked Queries}

When the query image is predicted as \texttt{unmasked}, the entire face is considered reliable for identity matching. Therefore, MGFace directly computes the cosine similarity between the query global feature $f_q$ and each gallery global feature $f_j$:

\begin{equation}
\mathbf{S}_{\text{global}}(q,\mathcal{G})
=
\left\{
\cos(f_q,f_j)
\right\}_{j=1}^{n},
\quad
\cos(f_q,f_j)
=
\frac{f_q^\top f_j}{\|f_q\|\|f_j\|}.
\end{equation}

The final ranked list is obtained by sorting all gallery images according to their global similarities:

\begin{equation}
\begin{aligned}
\mathcal{R}_{\text{global}}(q,\mathcal{G})
&=
\mathrm{Sort}
\left[
\mathbf{S}_{\text{global}}(q,\mathcal{G})
\right] \\
&=
\{g_{r_1},g_{r_2},\dots,g_{r_n}\}, \\
\text{s.t.}\quad
\cos(f_q,f_{r_1})
&\geq
\cos(f_q,f_{r_2})
\geq
\dots
\geq
\cos(f_q,f_{r_n}).
\end{aligned}
\end{equation}

This branch avoids unnecessary patch-level computation for normal face images, thereby improving efficiency.

\subsubsection{Two-Stage Mask-Aware Matching for Masked Queries}

When the query image is predicted as \texttt{masked}, the lower facial region may contain unreliable information due to occlusion. To improve robustness, MGFace performs matching in two stages: global candidate retrieval followed by mask-aware patch-level re-ranking.

\paragraph{Stage 1 - Global candidate retrieval}
MGFace first computes the global cosine similarity between the query and all gallery images. The top-$k$ most similar gallery images are then selected as candidate matches:

\begin{equation}
\begin{aligned}
\mathcal{G}_k(q,\mathcal{G})
&=
\mathrm{TopK}
\left[
\mathbf{S}_{\text{global}}(q,\mathcal{G})
\right] \\
&=
\mathcal{R}_{\text{global}}(q,\mathcal{G})_{[1:k]},
\quad k<n \\
&=
\{g_{r_1},g_{r_2},\dots,g_{r_k}\}.
\end{aligned}
\end{equation}

This stage provides an efficient coarse retrieval step and reduces the search space for subsequent patch-level re-ranking.

\paragraph{Stage 2 - Mask-aware patch-level re-ranking}
MGFace then re-ranks the top-$k$ candidates using patch-level features. Since face images are aligned using facial landmarks before feature extraction, their spatial layouts are relatively consistent. This allows MGFace to approximately localize occlusion-prone regions in the patch-level feature map.

For masked faces, the lower facial region is typically occluded, while the upper region, including the eyes and forehead, remains visible and informative for identity matching. Therefore, MGFace applies a mask-aware patch filter that retains only the upper spatial rows of the patch-level feature map. Given a patch-level feature map $P \in \mathbb{R}^{d_p \times N \times N}$, the filtered feature map is defined as:

\begin{equation}
\tilde{P}
=
P_{[:,\,1:M,\,1:N]}
\in
\mathbb{R}^{d_p \times M \times N},
\quad
M=\left\lfloor\frac{N+1}{2}\right\rfloor.
\end{equation}

\begin{figure}[h]
    \centering
    \includegraphics[width=0.85\linewidth]{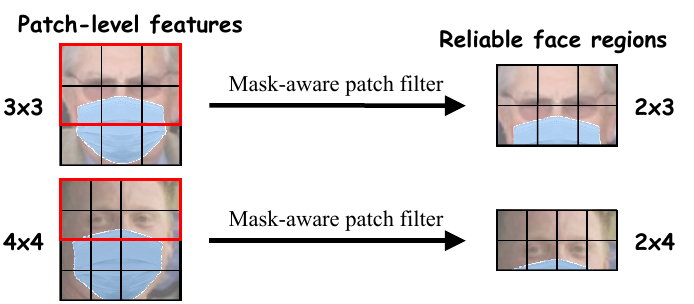}
    \caption{Illustration of the mask-aware patch filtering process. Only the upper facial patches are retained for masked face matching, while the lower patches affected by mask occlusion are discarded.}
    \label{fig:filter}
\end{figure}

This process is illustrated in Figure~\ref{fig:filter}. The cropped feature map is then flattened along the spatial dimensions into a sequence of $L$ patch vectors:

\begin{equation}
\tilde{P}
=
\{p^i \in \mathbb{R}^{d_p}\}_{i=1}^{L},
\quad
L=M\times N.
\end{equation}

For each candidate $g_{r_j} \in \mathcal{G}_k(q,\mathcal{G})$, MGFace computes the average cosine similarity between corresponding valid patch pairs after mask-aware filtering:

\begin{equation}
S_{\text{patch}}(q,g_{r_j})
=
\frac{1}{L}
\sum_{i=1}^{L}
\cos(p_q^i,p_{r_j}^i).
\end{equation}

The patch-level similarities for the top-$k$ candidates are denoted as
$\mathbf{S}_{\text{patch}}(q,\mathcal{G}_k)
=
\{S_{\text{patch}}(q,g_{r_j})\}_{j=1}^{k}$.
Finally, the top-$k$ candidates are re-ranked according to their patch-level similarities, while the remaining gallery images keep their original global-similarity order:

\begin{equation}
\mathcal{R}_{\text{patch}}(q,\mathcal{G})
=
\mathrm{Sort}
[
\mathbf{S}_{\text{patch}}(q,\mathcal{G}_k)
]
\oplus
\mathcal{R}_{\text{global}}(q,\mathcal{G})_{[k+1:n]},
\end{equation}

where $\oplus$ denotes list concatenation.

\section{Results}

\subsection{Settings and Datasets}
\subsubsection{Datasets}

We evaluate our method on the LFW-Mask dataset~\cite{emd}, which contains unmasked gallery images and synthetic masked query images generated from the original faces. To better reflect real-world scenarios, we extend the query set by combining both masked and unmasked images, while keeping the gallery set unchanged with only unmasked faces. During evaluation, duplicate images and pairs of synthetic masked images with their original counterparts are removed from the retrieval results. After retaining only identities with multiple images, the final extended dataset contains 1,659 identities. To train the mask classification head, we use the Mask Classifier dataset\footnote{\url{https://github.com/mahfujur1/Face-mask-Classification-PyTorch}}, which includes 2,757 training images and 1,276 validation images. The pretrained face recognition backbone is frozen, and only the mask classification head is optimized during training.
\begin{figure*}[h]
    \centering
    \includegraphics[width=\linewidth]{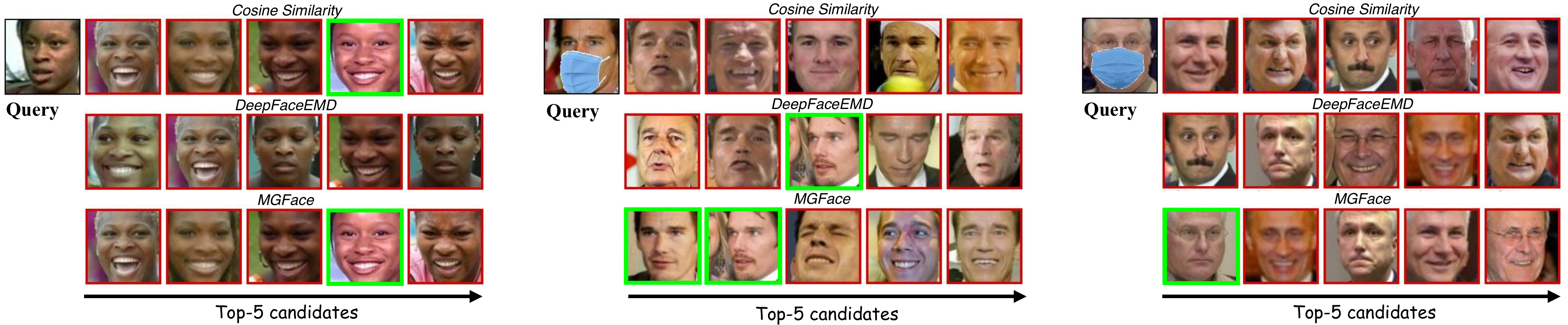}
    \caption{Qualitative top-\textit{5} retrieval results of cosine similarity, DeepFaceEMD, and MGFace. \textcolor{green}{Green} and \textcolor{red}{red} boxes denote correct and incorrect identity matches, respectively.}
    \label{fig:top5_visual}
\end{figure*}

\subsubsection{Settings}
To evaluate the effectiveness of the proposed method, we conduct experiments with different pretrained face recognition backbones, including FaceNet \cite{facenet} and ArcFace \cite{arcface}. Following the input requirements of each backbone, face images are resized to $128 \times 128$ for ArcFace and $160 \times 160$ for FaceNet. Unless otherwise stated, the number of candidates for the re-ranking stage is set to $k$=100. The default patch grid size is $3 \times 3$ for FaceNet and $4 \times 4$ for ArcFace. We adopt common face identification evaluation metrics, including Precision@1 (P@1), R-Precision (RP), and MAP@R (M@R), following prior works~\cite{emd,fastface}, to assess identification accuracy. In addition, we report the average query time to evaluate the computational efficiency of the proposed method.

\subsection{Face mask classification}

\begin{table}[h]
\centering
\caption{Mask classification accuracy after integrating the classification head into pretrained face recognition models.}
\label{tab:mask_cls}
\begin{tabular}{lccc}
\toprule
\multirow{2}{*}{\textbf{Model}} & \multicolumn{2}{c}{\textbf{Parameters}}           & \multirow{2}{*}{\textbf{Accuracy}} \\ \cmidrule{2-3}
                  & \textbf{Pretrained backbone} & \textbf{Mask head} &                                    \\ \midrule
FaceNet \cite{facenet}           & 27,910,327                   & 115,010            & 95.44\%                            \\
ArcFace \cite{arcface}          & 25,627,081                   & 33,090             & 97.91\%   \\              
\bottomrule
\end{tabular}%
\end{table}

Table~\ref{tab:mask_cls} reports the parameter cost and accuracy of the proposed mask classification head when integrated with pretrained face recognition backbones. Although the mask head is trained on the Mask Classifier dataset, it is evaluated on the \texttt{masked} and \texttt{unmasked} images from the extended LFW-Mask dataset to assess its generalization ability in the target evaluation setting. The results show that the mask head achieves high classification accuracy for both backbone settings, with 95.44\% for FaceNet and 97.91\% for ArcFace. Importantly, the number of additional parameters introduced by the mask head is small compared to those of the pretrained backbone. These results indicate that the proposed mask classification branch can provide reliable \texttt{masked}/\texttt{unmasked} predictions for mask-gated routing while adding only lightweight overhead to the original face recognition model.

\subsection{Face identification results}

\begin{table}[h]
\caption{Comparison of face identification performance on the extended LFW-Mask dataset using FaceNet and ArcFace}
\label{tab:main_results}
\resizebox{\columnwidth}{!}{%
\begin{tabular}{lcccccc}
\toprule
\multirow{2}{*}{\textbf{Method}} 
& \multicolumn{3}{c}{\textbf{FaceNet}}      
& \multicolumn{3}{c}{\textbf{ArcFace}}        \\ 
\cmidrule(lr){2-4}\cmidrule(lr){5-7} 
& \textbf{P@1} & \textbf{RP} & \textbf{M@R} 
& \textbf{P@1} & \textbf{RP} & \textbf{M@R} \\ 
\midrule
Cosine Similarity & 77.54 & 59.12 & \underline{55.84} & 87.22 & \underline{66.55} & \underline{64.11}        \\ \midrule
DeepFaceEMD$_{\text{APC}}$ & \underline{81.83} & \underline{59.63} & 55.17 & \underline{89.71} & 65.45 & 63.15        \\
DeepFaceEMD$_{\text{Uniform}}$ & 76.99 & 57.09 & 53.73 & 89.15 & 65.81 & 63.52        \\
DeepFaceEMD$_{\text{SC}}$ & 74.71 & 55.92 & 52.39 & 88.51 & 65.33 & 62.95        \\ \midrule 
\textbf{MGFace (ours)} & \textbf{82.92} & \textbf{60.91} & \textbf{58.05} & \textbf{91.91} & \textbf{68.57} & \textbf{66.81}        \\
\bottomrule
\end{tabular}%
}
\end{table}

Table~\ref{tab:main_results} reports the main face identification results using FaceNet and ArcFace as pretrained backbones. Overall, MGFace consistently achieves the best performance across all evaluation metrics and both backbones. Compared with conventional cosine similarity, MGFace improves P@1 by approximately 5\% on both FaceNet and ArcFace, while also achieving consistent gains in RP and M@R. Compared with DeepFaceEMD-based re-ranking methods, MGFace also obtains stronger results in most cases. On FaceNet and ArcFace, MGFace further improves all metrics over DeepFaceEMD$_{\text{APC}, \text{Uniform}, \text{SC}}$. These results suggest that the proposed mask-gated routing and mask-aware patch similarity are more effective for masked-face identification than applying general patch-wise re-ranking strategies. The routing strategy allows MGFace to choose the most suitable matching scheme based on the query conditions. It activates fine-grained patch-level matching only when masks are detected, concentrating on reliable facial regions. This approach enhances identification accuracy while minimizing unnecessary matching complexity for queries without masks.

To evaluate the computational efficiency of MGFace, we compare our conditional re-ranking strategy with DeepFaceEMD$_{\text{APC}}$ using the FaceNet backbone. As shown in Table~\ref{tab:efficient}, our proposed method reduces the total query time from 182.59s to 8.21s, corresponding to a speedup of over 20$\times$. In addition, MGFace reduces peak VRAM usage from 2.77 GB to 1.79 GB. These results demonstrate that MGFace provides a more efficient alternative to EMD-based re-ranking by avoiding unnecessary patch-level matching and using a lightweight mask-aware re-ranking strategy only when required.

\begin{table}[h]
\centering
\caption{Computational efficiency comparison between DeepFaceEMD and MGFace.}
\label{tab:efficient}
\begin{tabular}{lcc}
\toprule
\textbf{Method} & \textbf{Query time} & \textbf{VRAM allocated}  \\ \midrule
DeepFaceEMD \cite{emd}     & 182.59s                  & 2.77 GB                        \\
MGFace (ours)          & 8.21s                   & 1.79 GB                      \\ \bottomrule
\end{tabular}%
\end{table}

\subsection{Qualitative results}

Figure~\ref{fig:top5_visual} presents representative top-5 retrieval results of cosine similarity, DeepFaceEMD$_{\text{APC}}$, and MGFace. For \texttt{unmasked} queries, cosine similarity remains highly effective, motivating the mask-gated routing design of MGFace to avoid unnecessary re-ranking. For \texttt{masked} queries, re-ranking methods generally improve retrieval quality. At the same time, MGFace achieves more reliable results by concentrating patch-level matching on visible face regions. The results show that MGFace effectively adapts to both \texttt{masked} and \texttt{unmasked} queries, improving masked-face retrieval while preserving strong performance on unmasked faces through mask-gated routing.

\subsection{Ablation study}

\subsubsection{Different query sets}


We evaluate all methods with the FaceNet backbone under three query settings: \texttt{unmasked}, \texttt{masked}, and \texttt{unmasked} $\oplus$ \texttt{masked}. As shown in Table~\ref{tab:performance_transposed}, cosine similarity performs best on \texttt{unmasked} queries, while DeepFaceEMD$_{\text{APC}}$ slightly degrades performance, suggesting that patch-level re-ranking is unnecessary for fully visible faces. The proposed method maintains nearly the same performance as cosine similarity through mask-gated routing. For \texttt{masked} queries, MGFace achieves the best results by applying mask-aware patch-level matching on reliable facial regions. Overall, MGFace obtains the best performance on the combined query set, showing its effectiveness across different query conditions.

\begin{table}[h]
\centering
\caption{Comparison of face identification performance on different query sets.}
\label{tab:performance_transposed}
\begin{tabular}{llccc}
\toprule
\textbf{Query set}            & \textbf{Metric} & \textbf{Cosine} & \textbf{DeepFaceEMD$_{\text{APC}}$} & \textbf{MGFace} \\ \midrule
\multirow{3}{*}{\begin{tabular}[c]{@{}l@{}}\texttt{unmasked}\\ $\oplus$\\ \texttt{masked}\end{tabular}}    & P@1             & 77.54                      & \underline{81.83}                         & \textbf{82.92}  \\
                        & RP              & 59.12                      & \underline{59.63}                         & \textbf{60.91}  \\
                        & M@R             & \underline{55.84}                & 55.17                               & \textbf{58.05}  \\ \midrule
\multirow{3}{*}{\texttt{unmasked}}   & P@1             & \textbf{95.62}             & 94.76                               & {\underline{95.61}}     \\
                        & RP              & \textbf{74.73}             & 72.86                               & \underline{74.71}     \\
                        & M@R             & \textbf{73.61}             & 71.45                               & {\underline{73.58}}     \\ \midrule
\multirow{3}{*}{\texttt{masked}} & P@1             & 59.47                      & \underline{68.89}                         & \textbf{70.24}  \\
                        & RP              & 43.50                      & \underline{46.41}                         & \textbf{47.11}  \\
                        & M@R             & 38.08                      & \underline{41.49}                         & \textbf{42.51}  \\ \bottomrule
\end{tabular}%

\end{table}

\subsubsection{ Effect of different \textit{k}}



\begin{figure}[h]
    \centering
    \includegraphics[width=0.9\linewidth]{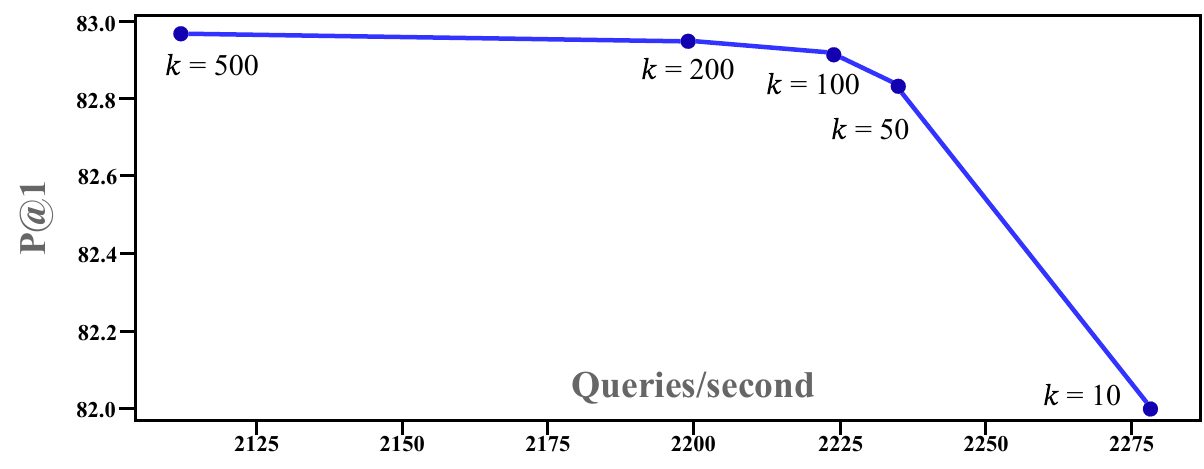}
    \caption{Accuracy-efficiency trade-off under different top-$k$ values in the re-ranking stage.}
    \label{fig:topk}
\end{figure}

We study the effect of the number of candidates $k$ used in the re-ranking stage. As shown in Figure~\ref{fig:topk}, increasing $k$ generally improves identification performance. Still, it reduces query speed due to the larger number of candidates involved in patch-level re-ranking. However, the performance gain becomes marginal when $k \geq 100$. In addition, P@1 increases only from 82.92 to 82.97 when $k$ increases from 100 to 500. At the same time, the query speed decreases from approximately 2224 to 2113 queries per second. Therefore, we set $k=100$ as the default value, which provides a good trade-off between accuracy and efficiency.

\subsubsection{Grid size selection for patch-level features}

We study the effect of different patch grid sizes. As shown in Table~\ref{tab:grid_size}, increasing the grid size from $4\times4$ to $8\times8$ slightly improves all metrics, suggesting that finer patch representations provide more informative local cues for mask-aware re-ranking. However, performance drops substantially at $16\times16$, likely because these features are extracted from shallower layers with weaker semantic representations. Therefore, we use the $4\times4$ setting as the default, which provides a good balance between accuracy and computational efficiency.

\begin{table}[h]
\centering
\caption{Effect of different grid sizes with the ArcFace backbone}
\label{tab:grid_size}
\begin{tabular}{lccc}
\toprule
\textbf{Grid size} & \textbf{P@1} & \textbf{RP} & \textbf{M@R} \\ \midrule
4$\times$4         & 91.91        & 68.57       & 66.81        \\
8$\times$8         & 92.30        & 68.75       & 67.04        \\
16$\times$16       & 81.84        & 58.08       & 55.05        \\ \bottomrule
\end{tabular}%
\end{table}

\section{Discussion and conclusion} \label{conclusion}

\subsection{Limitations}
Although MGFace improves the efficiency and robustness of masked-face identification, it has two main limitations. First, its conditional routing depends on the accuracy of the mask classification head; incorrect predictions may send a query to a suboptimal matching branch. Second, the current mask-aware filtering uses a fixed upper-face crop based on the assumption that masks occlude the lower facial region. This may be less effective for inaccurate face alignment or other irregular occlusions such as sunglasses. Future work will explore adaptive occlusion localization and learnable region selection to handle more diverse occlusion patterns.

\subsection{Conclusion}

We proposed MGFace, a mask-gated face identification pipeline that conditionally routes query images according to their predicted mask status. By applying global matching to unmasked queries and mask-aware patch-level re-ranking only to masked queries, MGFace improves masked-face identification while avoiding unnecessary computation. Experiments on the extended LFW-Mask dataset demonstrate that MGFace achieves better performance and lower computational cost than existing re-ranking-based methods. Future work will explore more adaptive occlusion localization strategies to handle diverse occlusion types beyond facial masks.

\section{ACKNOWLEDGMENT}
This research is funded by University of Information Technology-Vietnam National University of Ho Chi Minh city under grant number CS4-2026-80038.

{\small
\bibliographystyle{IEEEtran}
\bibliography{ref}

@inproceedings{emd,
  title={DeepFace-EMD: Re-ranking Using Patch-wise Earth Mover's Distance Improves Out-Of-Distribution Face Identification},
  author={Hai Phan, Anh Nguyen},
  booktitle={CVF Conference on Computer Vision and Pattern Recognition},
  year={2022}
}

@ARTICLE{deocc2,
  author={Ge, Shiming and Li, Chenyu and Zhao, Shengwei and Zeng, Dan},
  journal={IEEE Transactions on Circuits and Systems for Video Technology}, 
  title={Occluded Face Recognition in the Wild by Identity-Diversity Inpainting}, 
  year={2020},
  volume={30},
  number={10},
  pages={3387-3397},
  doi={10.1109/TCSVT.2020.2967754}}

@INPROCEEDINGS{face_class_mapr,
  author={Che, Quang-Huy and Le, Huu-Truyen and Ngo, Man-Dat and Tran, Hoang-Loc and Phan, Dinh-Duy},
  booktitle={2023 International Conference on Multimedia Analysis and Pattern Recognition (MAPR)}, 
  title={Intelligent Attendance System: Combining Fusion Setting with Robust Similarity Measure for Face Recognition}, 
  year={2023},
  volume={},
  number={},
  pages={1-6},
  doi={10.1109/MAPR59823.2023.10288710}}

@ARTICLE{deocc1,
  author={Cai, Jiancheng and Han, Hu and Cui, Jiyun and Chen, Jie and Liu, Li and Zhou, S. Kevin},
  journal={IEEE Transactions on Information Forensics and Security}, 
  title={Semi-Supervised Natural Face De-Occlusion}, 
  year={2021},
  volume={16},
  number={},
  pages={1044-1057},
  doi={10.1109/TIFS.2020.3023793}}

@article{mask_class,
title = {Real-time masked face classification and head pose estimation for RGB facial image via knowledge distillation},
journal = {Information Sciences},
year = {2022},
doi = {https://doi.org/10.1016/j.ins.2022.10.074},
author = {Chien Thai and Viet Tran and Minh Bui and Dat Nguyen and Huong Ninh and Hai Tran}
}

@article{fuse2,
title = {Enhancing person re-identification via Uncertainty Feature Fusion Method and Auto-weighted Measure Combination},
journal = {Knowledge-Based Systems},
volume = {307},
pages = {112737},
year = {2025},
issn = {0950-7051},
doi = {https://doi.org/10.1016/j.knosys.2024.112737},
author = {Quang-Huy Che and Le-Chuong Nguyen and Duc-Tuan Luu and Vinh-Tiep Nguyen}
}

@inproceedings{fuse1,
author = {Zhou, Mingrui and Chen, Xiangyu and Zhang, Weiming and Huang, Yu and Wang, Yaru},
title = {Improving Occluded Facial Recognition Accuracy by Integrating Large Language Model},
year = {2025},
isbn = {9798400713163},
publisher = {Association for Computing Machinery},
address = {New York, NY, USA},
doi = {10.1145/3746709.3746890},
booktitle = {Proceedings of the 2025 6th International Conference on Computer Information and Big Data Applications},
location = {
},
series = {CIBDA '25}
}

@InProceedings{occlu2,
author="Guo, Jianzhu
and Zhu, Xiangyu
and Lei, Zhen
and Li, Stan Z.",
editor="Zhou, Jie
and Wang, Yunhong
and Sun, Zhenan
and Jia, Zhenhong
and Feng, Jianjiang
and Shan, Shiguang
and Ubul, Kurban
and Guo, Zhenhua",
title="Face Synthesis for Eyeglass-Robust Face Recognition",
booktitle="Biometric Recognition",
year="2018",
publisher="Springer International Publishing",
address="Cham",
pages="275--284",
isbn="978-3-319-97909-0"
}

@INPROCEEDINGS {reidmulticlass,
author = { Che, Quang-Huy and Tran, Gia-Nghia },
booktitle = { 2023 International Conference on Advanced Computing and Analytics (ACOMPA) },
title = {Efficient Multi-Class Object Re-Identification Approach with A Unified Model },
year = {2023},
volume = {},
ISSN = {},
pages = {99-105},
doi = {10.1109/ACOMPA61072.2023.00025},
month =Nov}

@conference{kwf,
author={Quang{-}Huy Che and Le{-}Chuong Nguyen and Gia{-}Nghia Tran and Dinh{-}Duy Phan and Vinh{-}Tiep Nguyen},
title={A Re-Ranking Method Using K-Nearest Weighted Fusion for Person Re-Identification},
booktitle={Proceedings of the 14th International Conference on Pattern Recognition Applications and Methods - ICPRAM},
year={2025},
pages={79-90},
publisher={SciTePress},
organization={INSTICC},
doi={10.5220/0013176100003905},
isbn={978-989-758-730-6},
issn={2184-4313},
}

@INPROCEEDINGS{occlu1,
  author={Song, Lingxue and Gong, Dihong and Li, Zhifeng and Liu, Changsong and Liu, Wei},
  booktitle={CVF International Conference on Computer Vision}, 
  title={Occlusion Robust Face Recognition Based on Mask Learning With Pairwise Differential Siamese Network}, 
  year={2019},
  volume={},
  number={},
  doi={10.1109/ICCV.2019.00086}}

@article{face_detect,
title = {Face mask detection using deep learning: An approach to reduce risk of Coronavirus spread},
journal = {Journal of Biomedical Informatics},
volume = {120},
pages = {103848},
year = {2021},
issn = {1532-0464},
doi = {https://doi.org/10.1016/j.jbi.2021.103848},
author = {Shilpa Sethi and Mamta Kathuria and Trilok Kaushik}
}

@ARTICLE{counting_face,
  author={Nguyen, Khanh-Duy and Nguyen, Huy H. and Le, Trung-Nghia and Yamagishi, Junichi and Echizen, Isao},
  journal={IEEE Access}, 
  title={Analysis of Fine-Grained Counting Methods for Masked Face Counting: A Comparative Study}, 
  year={2024},
  volume={12},
  number={},
  pages={27426-27443},
  doi={10.1109/ACCESS.2024.3367593}}

@INPROCEEDINGS{vggdataset,
  author={Cao, Qiong and Shen, Li and Xie, Weidi and Parkhi, Omkar M. and Zisserman, Andrew},
  booktitle={13th IEEE International Conference on Automatic Face \& Gesture Recognition}, 
  title={A Dataset for Recognising Faces across Pose and Age}, 
  year={2018},
  doi={10.1109/FG.2018.00020}
}

@INPROCEEDINGS{agedb,
  author={Moschoglou, Stylianos and Papaioannou, Athanasios and Sagonas, Christos and Deng, Jiankang and Kotsia, Irene and Zafeiriou, Stefanos},
  booktitle={2017 IEEE Conference on Computer Vision and Pattern Recognition Workshops (CVPRW)}, 
  title={AgeDB: The First Manually Collected, In-the-Wild Age Database}, 
  year={2017},
  volume={},
  number={},
  pages={1997-2005},
  doi={10.1109/CVPRW.2017.250}}

@INPROCEEDINGS{fastface,
  author={Phan, Hai and Le, Cindy X. and Le, Vu and He, Yihui and Nguyen, Anh Totti},
  booktitle={2024 IEEE/CVF Winter Conference on Applications of Computer Vision (WACV)}, 
  title={Fast and Interpretable Face Identification for Out-Of-Distribution Data Using Vision Transformers}, 
  year={2024},
  volume={},
  number={},
  pages={6289-6299},
  doi={10.1109/WACV57701.2024.00618}}

@INPROCEEDINGS{transface,
  author={Dan, Jun and Liu, Yang and Xie, Haoyu and Deng, Jiankang and Xie, Haoran and Xie, Xuansong and Sun, Baigui},
  booktitle={International Conference on Computer Vision}, 
  title={TransFace: Calibrating Transformer Training for Face Recognition from a Data-Centric Perspective}, 
  year={2023},
  volume={},
  number={},
  doi={10.1109/ICCV51070.2023.01887}}

@INPROCEEDINGS{cosface,
  author={Wang, Hao and Wang, Yitong and Zhou, Zheng and Ji, Xing and Gong, Dihong and Zhou, Jingchao and Li, Zhifeng and Liu, Wei},
  booktitle={2018 CVF Conference on Computer Vision and Pattern Recognition}, 
  title={CosFace: Large Margin Cosine Loss for Deep Face Recognition}, 
  year={2018},
  volume={},
  number={},
  doi={10.1109/CVPR.2018.00552}}

@article{centerloss,
  title={A Comprehensive Study on Center Loss for Deep Face Recognition},
  
  author={Yandong Wen and Kaipeng Zhang and Zhifeng Li and Yu Qiao},
  journal={International Journal of Computer Vision},
  volume={127},
  pages={668--683},
  year={2019},
  doi={10.1007/s11263-018-01142-4}
}

@InProceedings{facenet,
author = {Schroff, Florian and Kalenichenko, Dmitry and Philbin, James},
title = {FaceNet: A Unified Embedding for Face Recognition and Clustering},
booktitle = {Proceedings of the IEEE Conference on Computer Vision and Pattern Recognition (CVPR)},
year = {2015}
}

@INPROCEEDINGS{arcface,
  author={Deng, Jiankang and Guo, Jia and Xue, Niannan and Zafeiriou, Stefanos},
  booktitle={2019 IEEE/CVF Conference on Computer Vision and Pattern Recognition (CVPR)}, 
  title={ArcFace: Additive Angular Margin Loss for Deep Face Recognition}, 
  year={2019},
  volume={},
  number={},
  doi={10.1109/CVPR.2019.00482}}
}

\end{document}